\documentclass[11pt]{article}

\usepackage[preprint]{acl}

 \usepackage{microtype}

\usepackage[english,bidi=default]{babel} 
\babelfont{rm}{TeXGyreTermesX} 
\usepackage{amsfonts}
\usepackage{amsmath}
\usepackage{amssymb}
\usepackage{booktabs}
\usepackage{url}
\usepackage{hyperref}
\usepackage{nicefrac}
\usepackage{subcaption}
\usepackage[inline]{enumitem}
\usepackage{amsmath}
\usepackage{graphicx}
\graphicspath{{figures/}}

\usepackage{enumitem}
\newlist{inlinelist}{enumerate*}{1}
\setlist[inlinelist,1]{label=(\roman*)}
\usepackage{linguex}
\title{Causal Interventions on Continuous Variables: A Case Study on Verb Bias in Steering Vectors for In-Context Learning}

\author{%
  Zhenghao Herbert Zhou  \hspace{.5in} R. Thomas McCoy  \hspace{.5in} Robert Frank \\
  Department of Linguistics\\
  Yale University\\
  New Haven, CT 06511, USA \\
  \texttt{\{herbert.zhou, tom.mccoy, robert.frank\}@yale.edu} \\
}

\begin{document}

\maketitle
\begin{abstract}
Causal interventions in language model representations have largely targeted discrete features, like grammatical  number.
However, language models must also make use of features that are graded.
We introduce a method for causal intervention on continuous variables: given activation vectors paired with a graded target variable, we localize a low-dimensional direction for that variable and use this direction to edit a vectors toward counterfactual target values. 
We apply this method to a continuous feature that is well-studied in psycholinguistics, namely verb bias (which reflects which syntactic structures tend to follow a given verb).
We show that verb bias is causally represented in steering vectors extracted from large language models: counterfactual edits to verb bias systematically shift downstream structural preferences. 
Verb bias has also previously been linked to in-context learning; in further analyses, we find that
steering vectors encode error signals that could drive the error-driven update behavior seen in in-context learning 
but that these aspects of the steering vectors are not causally used in downstream production.
Overall, these results show causal interventions can be applied to continuous variables, though connecting continuous variables to in-context learning remains a challenge. 
\end{abstract}

\section{Introduction}

Causal interpretability methods have emerged as a promising method for identifying internal representations and processing mechanisms in neural networks \citep{geiger2021causal,mueller2024quest}. 
Such work proceeds by localizing features in the representations of large language models (LLMs) and verifying their role in processing through causal intervention. 
Previous studies have focused on \textit{discrete} properties.
For instance, in work that focuses on the domain of language, features studied in this way include relative clause boundaries~\citep{ravfogel2021counterfactual},  subject number~\citep{hao2023verb}, nominal case~\citep{ozaki2025lstm}, the presence of a filler-gap dependency~\citep{boguraev2025causal}, and others. 

In this study, we extend the methods of causal intervention to \textit{continuous} properties.  
Many aspects of human language processing and conceptual knowledge are inherently graded. Expanding in this direction is thus important for enabling causal interventions to handle a broader range of features that might influence LLMs.
Specifically, given a set of activation vectors paired with a continuous target variable, our method aims to identify a low-dimensional direction corresponding to that variable and performs minimal edits to each vector toward specified counterfactual target values. 

We present a case study based on \textbf{verb bias}, a term which refers to the graded preference for a verb to occur with one syntactic structure compared to another that is semantically equivalent (e.g., \textit{I gave her a book} vs.\ \textit{I gave a book to her}). 
To operationalize the causal effect of verb bias, we leverage the psycholinguistic phenomenon of \textit{structural priming}, the tendency for speakers to reproduce recently encountered syntactic structures; the sentence whose structure gets reproduced is called the prime sentence (see Section~\ref{subsec:priming} for a review).
Structural priming is robustly found in both humans (e.g.,~\citealp{pickering1998representation}) and LLMs (e.g.,~\citealp{sinclair2022structural}). 
Because lexical properties of the verb present in a prime sentence impact the size of the priming effect, we can use changes in the amount of priming as an indicator of successful causal manipulation of verb bias.

A second motivation for studying verb bias in structural priming is its broader connection to \textit{in-context learning} (ICL), the ability for LLMs to adapt to novel tasks in context without weight updates.
\citet{zhou-etal-2025-context} proposes that structural priming should be conceptualized as a type of ICL: both involve an adaptation mechanism that depends on error-driven updates.
To simulate priming in LLMs, we use additive steering vectors, a compact way to capture ICL task representations (see Section~\ref{subsec:steering_vectors}).
We first show that steering vectors extracted from prime sentences induce standard structural priming.
We then use continuous variable intervention on steering vectors to show that a verb bias encoding that we identify causally determines priming magnitude, as counterfactual editing of verb bias causes predictable changes in downstream preferences for syntactic structures.

Additionally, we use the same intervention framework to ask what steering vectors capture about ICL. Prior work suggests that structural priming in humans and in LLMs involve an error-driven update: more surprising combinations of a verb and a syntactic structure produce larger priming effects. 
By causally editing an error-signal-like variable in steering vectors, we show that information about divergence from expectations is present and displays gradient update behavior under explicit intervention. However, we also find that steering-vectors alone do \textit{not} reproduce the error-driven priming behavior.
This suggests that the technique of additive steering captures an important component of ICL, but does not reproduce the full dynamic update that ICL accomplishes.

This work contributes to  interpretability, ICL, and psycholinguistics in several ways:

\ex. We present a method for performing causal interventions on continuous variables.

\ex. We manipulate the continuous verb bias in steering vectors and demonstrate its causal role in modulating models' downstream structural preference in priming.

\ex. We show that steering vectors simulate some aspects of structural priming as a form of ICL but do not fully capture the error-driven update process in ICL.

\ex. We propose a map between aspects of ICL and mechanisms in structural priming. 

\section{Background}


\subsection{Verb Bias and Structural Priming}\label{subsec:priming}

\textbf{Verb bias} refers to the preference for a verb to occur with one semantically equivalent structure rather than another. We focus here on the `dative alternation' between two structures associated with ditransitive verbs: the double object (DO) as in \textit{give} [$_{\mbox{\tiny NP}}$ \textit{the child}] [$_{\mbox{\tiny NP}}$ \textit{the ball}] and the preposition dative (PD) as in \textit{give} [$_{\mbox{\tiny NP}}$ \textit{the ball}] [$_{\mbox{\tiny PP}}$ \textit{to the child}]. The strength of the preference depends on the identity of the verb and varies continuously across verbs \citep{hawkins2020investigating}. As a result, the representation of verb bias must encode a continuous range of values. 

In psycholinguistics, the dative alternation is a well-studied domain for \textbf{structural priming}, the tendency for speakers or comprehenders to reuse a recently encountered syntactic structure~\citep{bock1986syntactic}. 
For instance, if a speaker hears a \textit{prime sentence} in PD, this increases her likelihood to produce a subsequent \textit{target sentence} in PD as opposed to DO.
Individual verbs exhibit a distinctive bias, a continuously variable relative preference for the two dative structures, and this determines the baseline to which the priming condition can be compared. 
An important aspect of structural priming is the \textit{inverse frequency effect} (IFE): less preferred prime structures tend to produce stronger priming effects than more preferred ones (\citealp{jaeger2008implicit,bernolet2010does,kaschak2011structural}).
This means that verb bias has a continuous effect on the priming magnitude: a PD prime containing a strongly DO-biased verb, or a DO prime containing a strongly PD-biased verb, should induce a larger priming effect than a prime that matches the verb’s baseline bias. 
In this study, we use structural priming to study causal intervention on verb bias representations in LLMs. 

\subsection{Priming as In-Context Learning}

While structural priming is usually viewed as a form of linguistic adaptation in humans~\citep{jaeger2013alignment}, an analogous phenomenon in LLMs is often discussed under the guise of \textbf{in-context learning} (ICL): a model’s ability to adapt its behavior based on a small number of demonstrations without parameter updates~\citep{brown2020language}. 
A central question in the study of ICL is what mechanism supports this in-context adaptation. 
Previous work has proposed that ICL may implicitly perform Bayesian updates~\citep{xie2021explanation} or gradient descent  \citep{pmlr-v202-von-oswald23a,dai2023metaoptimization}. 
These accounts differ in their mechanistic assumptions but share the idea that the model’s behavior on subsequent inputs is continuously adapted by the context in systematic ways.

Along this line of reasoning, \citet{zhou-etal-2025-context} use evidence from structural priming to argue that ICL involves an implicit error-driven update process.
The logic of this argument is parallel to error-based accounts of human structural priming: if a model updates more strongly in the presence of a larger error signal, then a larger priming effect should arise when there is a larger mismatch between the priming structure and the verb’s baseline expectation as determined by its bias. 
On this view, structural priming is a form of ICL, where the ``task'' is  defined as ``repeat the structure from the context''~\citep{chen-etal-2024-parallel}.
Thus, a prime sentence can be viewed as an in-context example: it provides evidence about a syntactic construction, and the model’s later structural preference should shift continuously in an error-driven way.


\subsection{Steering Vectors Capture Aspects of ICL}\label{subsec:steering_vectors}

One way to study the internal basis of ICL is through \textit{steering vectors} (also known as task vectors or function vectors), compact representations extracted from in-context demonstrations of tasks that are added to the model’s hidden states in a new inference run to induce corresponding task-dependent changes in downstream behaviors (e.g.,~\citealp{hendel2023context,todd2024function}). 
Prior work has shown that steering vectors capture relevant task information for a wide range of tasks such as
factual knowledge retrieval (e.g., country-capital mapping) and morphological inflections (e.g., past-present conversion).

These results suggest that at least some aspects of the mechanism underlying ICL are captured by  
such steering vectors. 
However, the fact that steering vectors can induce task-aligned behaviors does not imply that they capture the full mechanism by which ICL operates, 
leaving it unclear whether the addition of such steering vectors is fully equivalent to the original in-context computation.

In this study, we apply the steering vector approach to the phenomenon of structural priming.
We first test whether steering vectors encode the verb bias information that is crucial for inducing priming behaviors.
Moreover, we apply continuous variable editing to investigate to what extent the verb bias encodings that we have identified are causally involved in determining downstream structural preferences.
Next, to investigate whether steering vectors also capture the implicit error-driven update processes hypothesized for ICL, we further test whether an error-signal-like variable combining verb bias with observed prime structure is present in the same vectors and whether it is naturally deployed by additive steering. 


\begin{figure*}[t]
  \centering
  \includegraphics[width=0.93\textwidth]{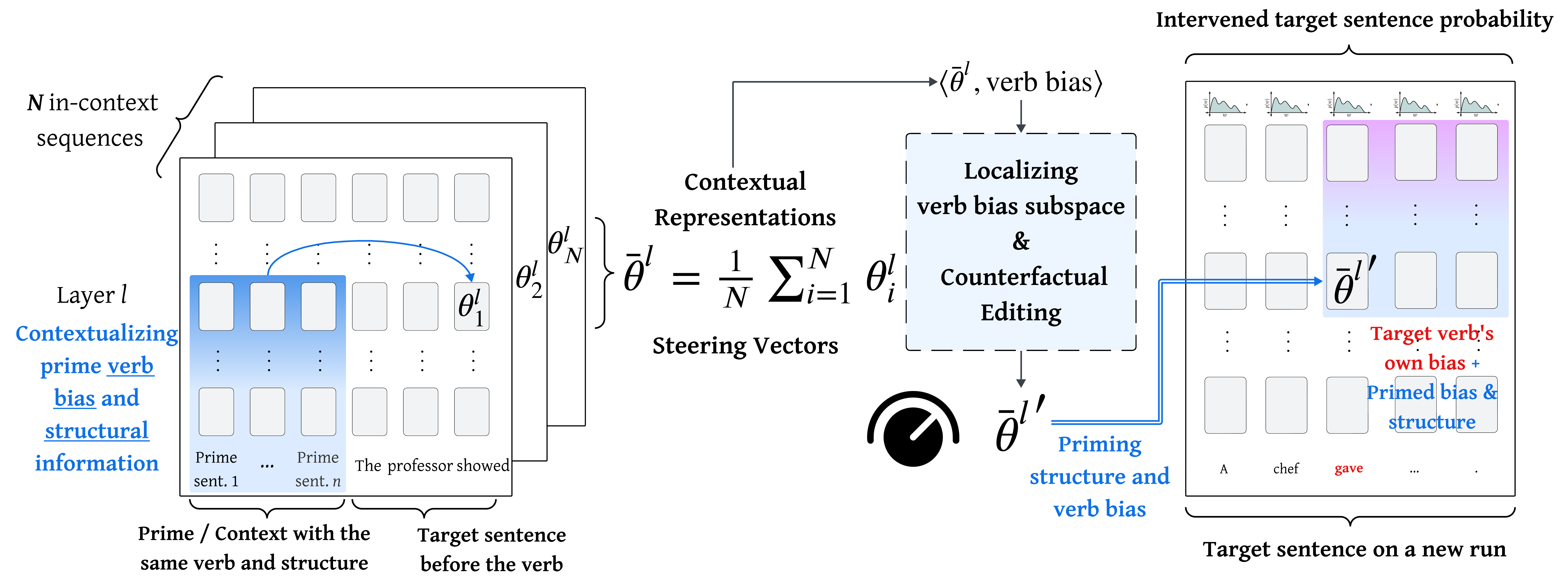}  
  \vspace{-.1in}
  \caption{An illustration of the procedure of extracting steering vectors in-context, applying continuous counterfactual editing, and injecting to the zero-shot target sentences. 
  }
  \label{fig:fv_extraction}
\end{figure*}

\section{Experiment 1: Steering Vectors from ICL Induce Structural Priming}\label{sec:exp1}

We first ask whether steering vectors extracted from priming contexts can induce structural priming effects. 
To preview our results, we find that they can, suggesting that they encode the continuous verb biases that are at the heart of structural priming; this finding lays the foundation for the experiments that follow,
which involve analyzing \textit{how} steering vectors encode continuous verb bias and then causally intervening on these encodings.


\subsection{Priming with Steering Vectors}\label{subsec:FV_extraction}

We use the dataset from \citet{zhou-etal-2025-context}, which is adapted from the Core Dative \textsc{Prime-LM} Corpus from \citet{sinclair2022structural}. The dataset contains 22 ditransitive verbs and 23,100 synthetically generated DO and PD sentences.
Following \citet{hendel2023context}, we define an ICL sequence as a concatenation of demonstrations (in our case, prime sentences) followed by a query (target sentence). As shown in Figure~\ref{fig:fv_extraction}, for a verb-structure combination $\langle v, s \rangle$, we sample $n$ sentences with verb $v$ and structure $s$ without lexical repetition for content words (except for the verb $v$) and with counterbalanced function words. The first $n-1$ sentences are concatenated and serve as the prime sentence context, and they are followed by the single remaining sentence that serves as the target sentence. For every layer $l$, we extract the hidden state $\theta^l$ at the target verb position, the point at which the model chooses between the PD and DO structural options. We average the $\theta^l$ obtained from 10 ICL sequences with the same $\langle v, s \rangle$ pair to obtain a set of steering vectors $\bar{\theta}_{\langle v, s \rangle}^l$. 
We hypothesize that a steering vector obtained in this way encodes abstract information about the ``implicitly defined task'', namely, producing the next sentence with structure $s$ influenced by verb $v$.
To simulate in-context priming effects, given a target sentence $t$, we inject steering vector $\bar{\theta}_{\langle v, s \rangle}^l$ to layer $l$ at the verb position of a target sentence presented without any priming context, by adding the steering vector to the corresponding original hidden state of $t$.

\begin{figure*}[t]
  \centering
  \includegraphics[width=0.91\textwidth]{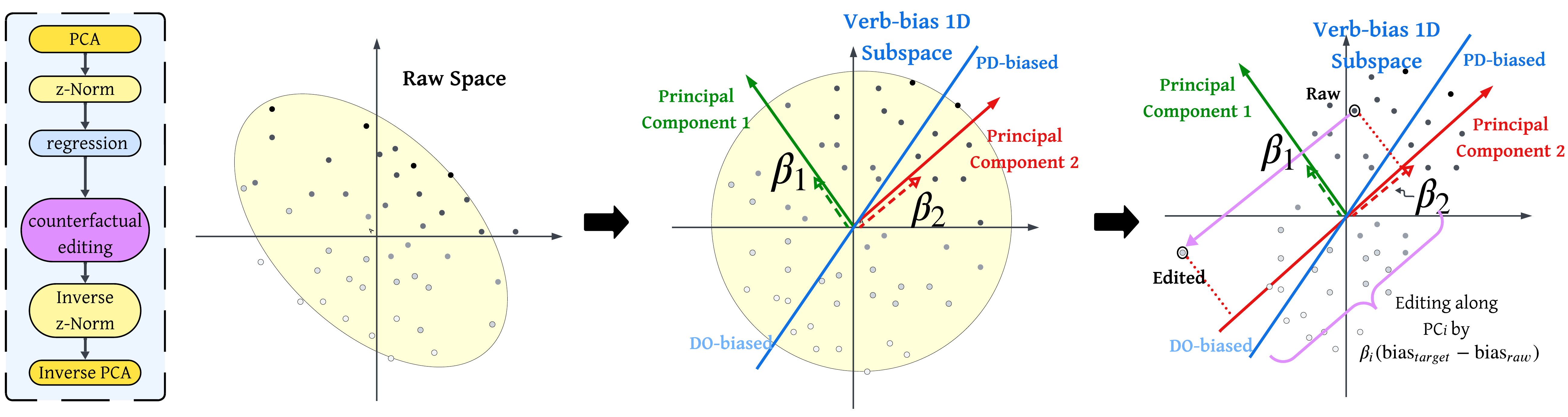} 
  \vspace{-.15in}
  \caption{An illustration of our continuous variable editing and intervention paradigm.}
  \label{fig:cont_int}
\end{figure*}

\subsection{Measuring Priming Effects}

We quantify the priming effect by measuring the direction and magnitude of shifts in the language model's structural preference. Following~\citet{zhou-etal-2025-context}, for each pair of target sentences $t_{\text{PD}}$ and $t_{\text{DO}}$, which differ only in their dative structure, we define the \textbf{raw PD preference} as: 
\begin{equation} \label{eq:raw_pref}
\text{RawPref}(t) = \frac{p(t_{\text{PD}})}{p(t_{\text{PD}}) + p(t_{\text{DO}})}
\end{equation}
where $p(t_{\text{PD}})$ is the probability of sentence $t_{\text{PD}}$ assigned by the model. 
A ratio closer to 1 means favoring the PD structure, and a ratio closer to 0 means favoring the DO structure.
We then define the \textbf{primed PD preference} by measuring the same quantity with the steering vector injected: 
\begin{equation} \label{eq:prime_pref}
\resizebox{0.48\textwidth}{!}{%
$\displaystyle
\text{PrimedPref}(t \vert \theta_{\langle v, s \rangle})
=
\frac{
p(t_{\text{PD}} \vert \theta_{\langle v, s \rangle})
}{
p(t_{\text{PD}} \vert \theta_{\langle v, s \rangle})
+
p(t_{\text{DO}} \vert \theta_{\langle v, s \rangle})
}
$%
}
\end{equation}
We say that steering vectors succeed in simulating the priming effect if they \textit{shift} the target sentence's structural preference towards the prime structure direction: for all verbs $v$, $\text{PrimedPref}(t \vert \theta_{\langle v, \text{PD} \rangle}) > \text{RawPref}(t)$ and $\text{PrimedPref}(t \vert \theta_{\langle v, \text{DO} \rangle}) < \text{RawPref}(t)$.

\subsection{Results}

\begin{figure}[t]
    \centering
    \includegraphics[width=0.45\textwidth]{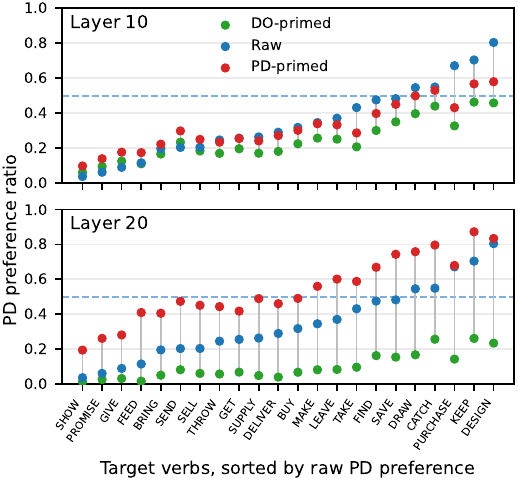}
     \vspace{-.15in}
    \caption{The raw and primed preference ratios for the 22 target verbs, with steering vectors extracted from layer 10 (top) and 20 (bottom).}
    \label{fig:exp1}
\end{figure}


For all experiments, we examined models from the Llama family~\citep{touvron2023llama}: Llama-{2-13B, 3-1B, 3-8B}. 
We show results for Llama-2-13B (40 layers in total, 5120 hidden dimensions) in the rest of the paper since we found no qualitative difference across the models.
We tested steering vectors extracted from layers \{5, 10, 15, 20, 30, 35\}, as high computational costs prevented us from studying all layers.
Following the procedure in Section~\ref{subsec:FV_extraction}, we extracted 50 steering vectors for all $\langle v, s\rangle$ prime combinations and injected them into a fixed set of 250 target pairs for each target verb.\footnote{To avoid measuring trivial lexical repetition effects, we exclude cases where the prime verb and target verb are identical, which could trigger a larger priming effect, also known as the lexical boost effect~\citep{pickering1998representation}.}

We find that steering vectors induce structural priming effects in mid-to-late layers, but not early layers. 
As shown in Figure~\ref{fig:exp1} (top), steering vectors extracted from layer 10 do not reliably produce structure-congruent priming: for most target verbs, both DO-based and PD-based steering vectors can shift the model in the same direction.
In contrast, from layer 15 onward, steering vector injection shows the expected priming effect: for all verbs, vectors extracted from PD primes increase the model’s PD preference, while vectors from DO primes decrease it, as shown for layer 20 in Figure~\ref{fig:exp1} (bottom). 
This establishes that steering vectors have a causal effect on a model’s downstream structural preferences, a result that establishes the behavioral foundation for the rest of this paper. 




\begin{figure*}[t]
    \centering
    \includegraphics[width=0.44\textwidth]{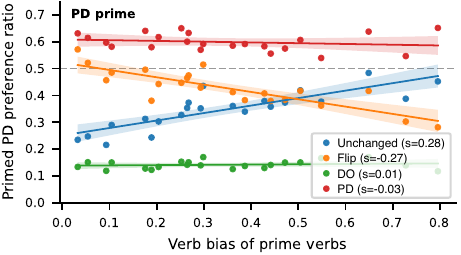}
    \hfill
    \includegraphics[width=0.44\textwidth]{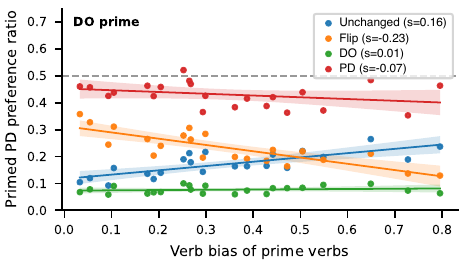}
    \vspace{-.14in}
    \caption{
    Interventions on verb bias in steering vectors with PD primes (left) and DO primes (right).
    Each point shows the primed PD preference ratio for a prime verb, with fitted regression lines and 95\% confidence intervals.
    }
    \label{fig:exp2_static_bias_intervention}
\end{figure*}

\section{Experiment 2: Verb Bias Causally Modulates Structural Preference}\label{sec:exp2}

Results from Experiment 1 indicate that steering vectors encode abstract, structural 
information that impacts a model's syntactic choices. 
We now ask how this effect is modulated in a continuous way. 

\subsection{Methodology}\label{sec:method}

We present a procedure for localizing and causally intervening on continuous variables in vector representations. 
Given a set of vectors $h_i \in \mathbb{R}^d$, each paired with a continuous scalar label $y_i$, our goal is  not only to test whether $y$ is predictable from $h$, but also to construct a counterfactual vector $h_i'$ such that the model’s predicted value of the target variable is changed to a specified value $y_i^*$ with minimal change in the original vector $h_i$.

\paragraph{Localizing Continuous Variables}

We first map the vectors into a lower-dimensional space (Figure~\ref{fig:cont_int}, left and middle). 
For each vector $h_i$, we apply PCA to obtain $z_i = \text{PCA}(h_i) \in \mathbb{R}^k$ with $k \ll d$.
We then normalize each PCA dimension to have zero mean and unit variance: $\tilde{z}_{ij} = \frac{z_{ij} - \mu_j}{\sigma_j}$.
On these normalized coordinates, we fit a regression model that predicts the continuous target variable. For a variable with range in $\mathbb{R}$, we use $\hat{y_i} = \beta_0 + \beta^\top\tilde{z}_i$.
For bounded variables in the range $(0, 1)$, we use a generalized linear model with an appropriate link function $g$, such that $\hat{\eta_i} = g(\hat{y_i}) = \beta_0 + \beta^\top\tilde{z}_i$.
The experiments below use beta regression for the bounded verb bias variable and ordinary least squares (OLS) for signed continuous variables. In both cases, the fitted regression defines a direction vector $\beta$ in normalized PCA space along which the predicted target value changes.

\paragraph{Counterfactual Editing}

Given a vector $\tilde{z}_i$, the continuous variable's value predicted by our fitted model is $\hat{\eta_i} = \beta_0 + \beta^\top\tilde{z}_i$. 
For a desired counterfactual target value $y_i^*$, we compute the corresponding target on the model’s prediction scale, which is $\eta_i^* = y_i^*$ for OLS regression and $\eta_i^* = g(y_i^*)$ for beta regression.
To solve for a small update $\Delta \tilde{z}_i$ on the normalized vector such that $\eta_i^* = \beta_0 + \beta^\top (\tilde{z}_i + \Delta \tilde{z}_i)$, we apply the minimal $L_2$-norm update specified by the regression direction, given by $\Delta \tilde{z}_i = \frac{\eta_i^* - \eta_i}{\Vert \beta_S \Vert^2}\beta_S$, where $S$ denotes the set of dimensions selected to be edited, and $\beta_S$ is the coefficient vector restricted to those dimensions, illustrated in the right panel of Figure~\ref{fig:cont_int}.
Coordinates outside $S$ are left unchanged. 
In practice, $S$ is selected based on the magnitude of regression coefficients: principle components (PCs) with higher coefficients are considered the most influential in perturbing the variable.\footnote{We also applied Lasso regression over $\beta_0$ and $\beta$ as a regularization method to filter out noise-like, uninfluential PCs.}
The edited vector in normalized PCA space is therefore $\tilde{z}_i' = \tilde{z}_i + \alpha \Delta \tilde{z}_i$, where $\alpha$ is a hyperparameter that controls the amount of editing. 
When $\alpha=1$, the edit moves the vector to the target value predicted by the fitted readout; smaller or larger values of $\alpha$ allow weaker or stronger interventions.
Finally, we invert the normalization and PCA transformation to map the edited vector back to the original representation space: $h_i' = \text{PCA}^{-1}(\sigma \tilde{z}_i' + \mu)$.
The edited vector can then be used in place of the original vector in the language model's forward pass to observe the downstream effect of our intervention.
See Appendix~\ref{app:leace} for the relation between the current method and prior concept erasure methods such as INLP ~\citep{ravfogel2020null} and LEACE~\citep{belrose2023leace}.

\subsection{Counterfactual Bias Editing}

For each prime verb $v$, we computed its \textbf{raw verb bias} $b_v \in (0, 1)$ by averaging the $\text{RawPref}$ values over all sentence pairs with verb $v$ in the entire dataset. To test whether verb bias is represented in steering vectors, we fit a beta regression model from normalized PCA coordinates of 2200 steering vectors (keeping 50 principle components).

The beta regression identifies a 1-dimensional subspace in normalized PCA space that encodes verb bias information.
To investigate whether this verb bias subspace is causally implicated in LLM processing, we 
create counterfactual function vectors by scaling the verb bias values within the identified subspace using the method in Section~\ref{sec:method}. 
Given the continuous nature of verb bias, we consider the following conditions:
\begin{enumerate*}[label=(\arabic*)]
    \item \texttt{Unchanged}: the control condition keeping the original verb bias information, thus representing the original priming effects of the steering vectors;
    \item \texttt{Flip}: given a verb with bias $b_v$, modify it to $1-b_v$, thereby asking whether the graded relation between verb bias and priming effects can be reversed;
    \item \texttt{Fully Biased}: for all verbs, modify their biases to $1$ (fully \texttt{PD-biased}) or $0$ (fully \texttt{DO-biased}), eliminating the difference in bias across verbs.
\end{enumerate*}\footnote{We set the editing strength parameter $\alpha=1$ to ensure exact edits to the target verb bias values. We report results with $S=5$, i.e., the most influential 5 PCs. In practice, we found no significant difference in intervention effect when the selected dimensions are \{top 1 PC, top 5 PCs, and all PCs\}.}
If verb bias is causally involved in the steering vector’s downstream effect, then we predict that the priming magnitudes of \texttt{Flip} and \texttt{Unchanged} will be inversely correlated, while there should be no difference across verbs with respect to priming magnitudes in the two \texttt{Fully Biased} conditions. 

\subsection{Results}

\paragraph{Quantifying Intervention Effects}
Since the priming magnitude is modulated by the prime verb's bias, we quantify the priming effect induced by prime verbs.
For each prime verb $v$ and structure $s$, we compute the mean $\text{PrimedPref}(t \vert \theta_{\langle v, s \rangle})$ across all target sentences $t$ with all target verbs except for $v$.
For each of the 4 intervention conditions, we then fit a line of priming effect of each prime verb against its raw verb bias to show the correlation between verb bias and priming magnitude. 

\paragraph{Counterfactual Editing Results}
The positive slope for the \texttt{Unchanged} condition and negative slope for the \texttt{Flip} condition (Figure~\ref{fig:exp2_static_bias_intervention}) suggest that the correlation between verb bias and priming magnitude is effectively inverted through our manipulation of the counterfactual function vectors encoding verb bias. 
The two essentially flat slopes observed for the two \texttt{Fully Biased} (\texttt{DO} and \texttt{PD}) conditions suggest that when the relevant subspace of the counterfactual steering vectors is saturated, existing difference in verb biases are eliminated across individual prime verbs.
PD primes (the left panel of Figure~\ref{fig:exp2_static_bias_intervention}) and DO primes (the right panel) share the same slope pattern, while the intercept of PD primes is significantly higher than that of DO primes, indicating that the standard priming effect from Experiment 1 continues to be observed. 

\paragraph{Layerwise Dynamics}
The effect of verb bias intervention is visible even in early layers (starting at layer 5; see Appendix~\ref{app:layerwise}), before steering vectors reliably produce priming. 
This contrasts with the layerwise pattern in Experiment 1, where expected priming emerges only in mid-to-late layers. 
The dissociation suggests that verb bias is available as an early lexical variable, while the \textit{integration} of this bias with the actually observed prime structure emerges later; this integration process is a central focus in Experiment 3.


\paragraph{Interpretation}
The positive slope in the \texttt{Unchanged} condition also clarifies the role of verb bias on the priming effect induced by steering vectors.
They behave as an additive structural preference state: a highly PD-biased verb causes a larger shift to downstream PD preference than a less PD-biased verb.
In sum, continuous causal manipulation in the identified verb bias subspace shows the predicted downstream effect on priming magnitude.
It provides a central demonstration of our method: a continuous psycholinguistic variable can be localized in activation-derived vectors, counterfactually edited, and shown to have a predictable causal effect on model behavior.

\section{Experiment 3: Error Signal Editing Reveals a Limit of Steering Vectors}\label{sec:exp3}

\begin{figure}[t]
  \centering
  \includegraphics[width=0.79\columnwidth]{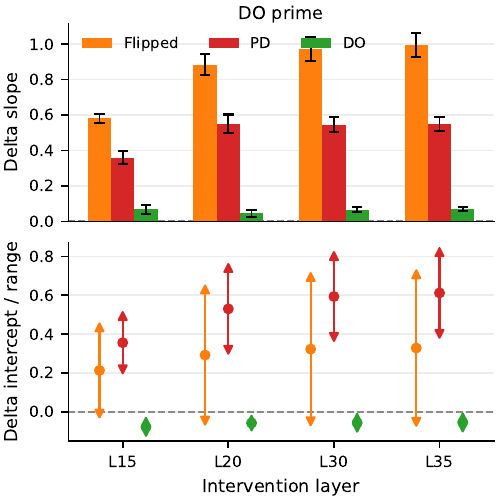} 
  \vspace{-.16in}
  \caption{The slopes (top) and ranges (bottom) of the $\Delta \text{Pref}$ metric for the three causal intervention conditions relative to the \texttt{Unchanged} baseline.}
  \label{fig:exp3}
\end{figure}

Experiment 2 shows that verb bias is causally editable in steering vectors. 
However, verb bias alone does not capture the full \textit{adaptation} process hypothesized in \textit{error-driven learning} account of priming (see Section~\ref{subsec:priming} and Appendix~\ref{app:priming_theory}). 
Under such accounts, priming magnitude is modulated by the \textit{gradient mismatch} between the observed prime structure and the expectation from the verb bias of the verb in the prime sentence. 
One key signature of this gradient mismatch is the inverse frequency effect (IFE) mentioned above: prime sentences involving less expected combinations of a verb and a syntactic structure have a larger gradient mismatch and thus produce stronger priming.
\citet{zhou-etal-2025-context} showed that for in-context priming, LLMs show the IFE. Since the IFE is a signature of error-driven learning, this finding provided evidence that language model priming (and perhaps ICL more broadly) can be viewed as implicitly involving error-driven learning.
In the current setup, the IFE predicts negative slopes instead of positive ones in the \texttt{Unchanged} condition in Figure~\ref{fig:exp2_static_bias_intervention}.
Thus, despite showing standard priming effects, steering vectors extracted under the current setup do not produce the IFE, thereby failing to capture the crucial error-driven nature of ICL.

To better understand this limitation, we ask whether steering vectors contain an error-signal-like variable, a crucial ingredient of error-driven updates, that combines the prime verb’s bias with the actually observed prime structure. 
For a prime verb with PD bias $b_v$ we define signed error as $e = b_v - s \in (-1,1)$, where $s=1$ for a PD prime and $s=0$ for a DO prime. 
Under this convention, positive values correspond to an update toward DO, and negative values correspond to an update toward PD. 
This variable is not intended to be a definitive estimate of the model’s internal prediction error; rather, it provides a theory-driven continuous target for testing whether the ingredients of error-driven adaptation are present and causally accessible in the steering-vector space.

\subsection{Causal Editing of Error Signals}

Following Experiment 2, we construct counterfactual steering vectors by editing the error signal value in the same 4 conditions: \texttt{Unchanged} (preserving the original steering vectors); \texttt{Flip} (reversing the sign of the error signal, asking whether a vector associated with an update in one direction can be transformed into one favoring the opposite direction); and the two \texttt{Fully Biased} conditions (setting $e'=-1$ for \texttt{PD} and $e'=1$ for \texttt{DO}).

Since the \texttt{Unchanged} slopes already reflect the original priming effect of the steering vectors, to study the effects of counterfactual editing, we analyze error signal edits relative to this baseline. 
For each prime verb, we compute $\Delta \text{Pref}_{\text{edited}} = \text{PrimedPref}_{\text{edited}} - \text{PrimedPref}_{\text{Unchanged}}$ for the 3 editing conditions.
This delta measure isolates the \textit{additional} effect of the error signal edit beyond the original steering vectors. We then ask whether $\Delta \text{Pref}$ varies systematically with the error signal variable in an error-driven way: do larger error signals cause larger updates on the downstream structural preference in the primed direction?
Under this setup, we predict that $\text{slope}(\Delta \text{Pref}_{\text{flip}}) > \text{slope}(\Delta \text{Pref}_{\text{PD}}) > \text{slope}(\Delta \text{Pref}_{\text{DO}}) > 0$ and $\text{intercept}(\Delta \text{Pref}_{\text{PD}}) > \text{intercept}(\Delta \text{Pref}_{\text{flip}}) \approx 0 > \text{intercept}(\Delta \text{Pref}_{\text{DO}})$ (see Appendix~\ref{app:error_signal} for details).

\subsection{Results}

As shown in Figure~\ref{fig:exp3}, error signal editing produces systematic, gradient shifts in downstream structural preference, confirming the predictions derived from an error-driven updating process.
Flipping the error signals produces a counter-bias shift whose magnitude varies with the prime verb’s baseline structural preference, relative to the \texttt{DO} and \texttt{PD} editing that eliminate differences across verbs.

However, this pattern holds only for layer 15 and onward but not earlier layers (Appendix~\ref{app:layerwise}), where error signal intervention has no significant effect despite the fact that prime structure is already linearly decodable (Appendix~\ref{app:diagnostics}).
This effect is consistent with results in Experiment 1, where steering vectors only induce structure-congruent priming starting at layer 15.
The distinction between Experiments 2 and 3 suggests that, although early layers contain separable ingredients of error-driven adaptation (verb bias and prime structure), further processing is required for them to be \textit{integrated} into an error signal that can be manipulated to demonstrate gradient-like updating behaviors.


In sum, we have showed that although the steering vectors do not produce the IFE that is the evidence for an error-driven updating process occurring in ICL, the error signal information is present in the vectors and can be manipulated to display the gradient error-driven updates.
This result reveals a limitation of additive steering vectors: they capture some aspects of in-context priming, but not the full dynamic update implemented by ICL.


\section{Discussion and Conclusion}\label{sec:discussion}


In contrast to prior work focused on discrete features,
our results show that cognitively-relevant continuous variables can be localized and edited in activation space: verb bias is linearly predictable from steering vectors, and counterfactual edits  produce systematic changes in downstream structural preference. This suggests that continuous variable intervention can serve not only as a diagnostic tool for asking whether a feature is encoded, but also as a causal tool for testing how graded representational information is used by models.

This approach complements recent work arguing that ICL involves learning or organizing concept subspaces~\citep{tang2026context}. 
In our setting, PCA and regression provide a simple way to identify behaviorally meaningful directions within the vector space, while counterfactual editing tests whether movement along these directions has the predicted behavioral consequences.  

\paragraph{Strengths and limitations of steering vectors}

Our results give a mixed picture of steering vectors as encodings of in-context adaptation. 
On the one hand, steering vectors extracted from prime sentences induce standard structural priming: PD-prime vectors increase target PD preference, while DO-prime vectors decrease it. 
Verb bias is recoverable from these vectors, and editing this variable causally changes the direction and magnitude of the structural preference. 
These findings support the view that steering vectors are an important part of the representational state induced by in-context priming.
Yet steering vectors do not capture the full error-driven character of in-context learning: unedited steering-vector injection produces standard priming but not the IFE. 
This suggests that the extracted vectors contain information about structural preference, but do not implement the update rule that determines how strongly a model should adapt as a function of prediction error. 
Our error signal analysis sharpens this point: error signal is present in the vectors and can be causally manipulated, but error-driven behavior does not naturally result from simple vector injection. 
This distinction between information present in a model's representation and causally used~\citep{lepori2026language} connects to recent work on the limitations of steering vectors in ICL~\citep{dong2025understanding}.

\paragraph{Connecting psycholinguistics and ICL}

In human sentence processing, two families of mechanisms have been used to explain structural priming. 
\textit{Transient activation}~\citep{pickering1998representation} treats priming as the residual activation of recently used structures, while \textit{implicit learning}~\citep{chang2006becoming} treats priming as an error-driven update that is stronger when the prime is less expected. 
Our steering-vector intervention resembles the former: adding a vector extracted from the prime shifts the model's structural expectations. 
However, the absence of the IFE under simple vector injection suggests that this operation does not capture the full implicit learning mechanism.

This distinction aligns with recent theoretical accounts of ICL where the effect of context is decomposed into steering-vector-like changes in activation states and implicit low-rank updates in MLP~\citep{dherin2025learning}. 
From this perspective, steering vectors may approximate the transient activation component of structural priming, while the error-driven component may require changes that are better characterized as updates to the computation rather than a single additive vector. 
Illustrated in Figure~\ref{fig:implication}, the present results therefore suggest a mechanistic level mapping between psycholinguistic mechanisms and mechanistic hypotheses about ICL: activation steering captures context-induced representational states, but error-driven adaptation may require a richer mechanism like low-rank updates. 
This mapping highlights how linguistic phenomena such as structural priming can shed light on parallel mechanisms in LLMs.

\section*{Limitations}

\paragraph{Scope of the steering-vector result}
Our result on the absence of error-driven adaptation effects applies only to the steering vectors extracted with our current procedure. 
We do not claim that all steering-vector-based methods, or activation steering more generally, are necessarily incapable of capturing error-driven components of ICL. 
Different extraction procedures, different target positions, or multi-vector interventions may recover aspects of the computation that are not captured here.

\paragraph{Position specificity of current steering vectors}
We extract steering vectors only from the target verb position. 
This choice is well motivated for studying static verb bias and the interaction between verb identity and prime structure at the verb position, but the relevant error-driven signal may emerge later in the sentence. 
Future work should compare vectors extracted from multiple positions, including the token position after the first object NP (i.e., the position that distinguishes the presence or absence of a preposition, which definitively determines the sentence structure) or the end of the prime sentence, to better localize where different components of priming are represented.

\paragraph{Linear assumptions about the representation of continuous variables}
Our intervention method relies on linear readouts in a low-dimensional subspace. 
This makes the method interpretable and tractable, but it may miss variables encoded nonlinearly or distributed across dimensions that are not well captured by the selected principal components. 
Relatedly, PCA is used here as a practical regularization step, not as a theoretically necessary component of continuous-variable intervention. 
Future work should more systematically compare PCA-based editing with closed-form concept-erasure and editing methods such as LEACE, and clarify when dimensionality reduction is helpful or harmful.

\paragraph{Interpretation of the error signal results}
The error signal intervention results should be interpreted cautiously. 
The diagnostics suggest that signed-error-like information is present in the steering-vector space, and the intervention results show that manipulating this dimension produces systematic downstream effects. 
However, this does not amount to a full mechanistic account of error-driven learning. In particular, the signed error signal variable is partly dominated by prime-structure information, and its downstream effects require a more complex interpretation than the verb bias intervention.


\bibliography{function_vector}

\begin{thebibliography}{31}
\providecommand{\natexlab}[1]{#1}

\bibitem[{Belrose et~al.(2023)Belrose, Schneider-Joseph, Ravfogel, Cotterell, Raff, and Biderman}]{belrose2023leace}
Nora Belrose, David Schneider-Joseph, Shauli Ravfogel, Ryan Cotterell, Edward Raff, and Stella Biderman. 2023.
\newblock {LEACE}: {Perfect} linear concept erasure in closed form.
\newblock \emph{Advances in Neural Information Processing Systems}, 36:66044--66063.

\bibitem[{Bernolet and Hartsuiker(2010)}]{bernolet2010does}
Sarah Bernolet and Robert~J. Hartsuiker. 2010.
\newblock Does verb bias modulate syntactic priming?
\newblock \emph{Cognition}, 114(3):455--461.

\bibitem[{Bock(1986)}]{bock1986syntactic}
J.~Kathryn Bock. 1986.
\newblock Syntactic persistence in language production.
\newblock \emph{Cognitive Psychology}, 18(3):355--387.

\bibitem[{Boguraev et~al.(2025)Boguraev, Potts, and Mahowald}]{boguraev2025causal}
Sasha Boguraev, Christopher Potts, and Kyle Mahowald. 2025.
\newblock Causal {Interventions} {Reveal} {Shared} {Structure} {Across} {English} {Filler}-{Gap} {Constructions}.
\newblock In \emph{Proceedings of the 2025 Conference on Empirical Methods in Natural Language Processing}, pages 25032--25053.

\bibitem[{Brown et~al.(2020)Brown, Mann, Ryder, Subbiah, Kaplan, Dhariwal, Neelakantan, Shyam, Sastry, Askell, Agarwal, Herbert-Voss, Krueger, Henighan, Child, Ramesh, Ziegler, Wu, Winter, Hesse, Chen, Sigler, Litwin, Gray, Chess, Clark, Berner, McCandlish, Radford, Sutskever, and Amodei}]{brown2020language}
Tom~B. Brown, Benjamin Mann, Nick Ryder, Melanie Subbiah, Jared Kaplan, Prafulla Dhariwal, Arvind Neelakantan, Pranav Shyam, Girish Sastry, Amanda Askell, Sandhini Agarwal, Ariel Herbert-Voss, Gretchen Krueger, Tom Henighan, Rewon Child, Aditya Ramesh, Daniel~M. Ziegler, Jeffrey Wu, Clemens Winter, and 12 others. 2020.
\newblock Language models are few-shot learners.
\newblock \emph{Advances in Neural Information Processing Systems}, 33:1877--1901.

\bibitem[{Chang et~al.(2006)Chang, Dell, and Bock}]{chang2006becoming}
Franklin Chang, Gary~S. Dell, and J.~Kathryn Bock. 2006.
\newblock Becoming syntactic.
\newblock \emph{Psychological Review}, 113(2):234.

\bibitem[{Chen et~al.(2024)Chen, Zhao, Yu, McKeown, and He}]{chen-etal-2024-parallel}
Yanda Chen, Chen Zhao, Zhou Yu, Kathleen McKeown, and He~He. 2024.
\newblock \href {https://doi.org/10.18653/v1/2024.acl-long.465} {Parallel {Structures} in {Pre}-training {Data} {Yield} {In}-{Context} {Learning}}.
\newblock In \emph{Proceedings of the 62nd Annual Meeting of the Association for Computational Linguistics (Volume 1: Long Papers)}, pages 8582--8592, Bangkok, Thailand. Association for Computational Linguistics.

\bibitem[{Dai et~al.(2023)Dai, Sun, Dong, Hao, Ma, Sui, and Wei}]{dai2023metaoptimization}
Damai Dai, Yutao Sun, Li~Dong, Yaru Hao, Shuming Ma, Zhifang Sui, and Furu Wei. 2023.
\newblock Why can {GPT} {Learn} {In}-{Context}? language {Models} {Implicitly} {Perform} {Gradient} {Descent} as {Meta}-{Optimizers}.
\newblock In \emph{ICLR 2023 Workshop on Mathematical and Empirical Understanding of Foundation Models}.

\bibitem[{Dherin et~al.(2025)Dherin, Munn, Mazzawi, Wunder, and Gonzalvo}]{dherin2025learning}
Benoit Dherin, Michael Munn, Hanna Mazzawi, Michael Wunder, and Javier Gonzalvo. 2025.
\newblock Learning without training: {The} implicit dynamics of in-context learning.
\newblock \emph{arXiv preprint arXiv:2507.16003}.

\bibitem[{Dong et~al.(2025)Dong, Jiang, Zhu, and Ning}]{dong2025understanding}
Yuxin Dong, Jiachen Jiang, Zhihui Zhu, and Xia Ning. 2025.
\newblock Understanding task vectors in in-context learning: {Emergence}, functionality, and limitations.
\newblock \emph{arXiv preprint arXiv:2506.09048}.

\bibitem[{Geiger et~al.(2021)Geiger, Lu, Icard, and Potts}]{geiger2021causal}
Atticus Geiger, Hanson Lu, Thomas Icard, and Christopher Potts. 2021.
\newblock Causal abstractions of neural networks.
\newblock \emph{Advances in neural information processing systems}, 34:9574--9586.

\bibitem[{Hao and Linzen(2023)}]{hao2023verb}
Sophie Hao and Tal Linzen. 2023.
\newblock Verb conjugation in transformers is determined by linear encodings of subject number.
\newblock In \emph{Findings of the Association for Computational Linguistics: EMNLP 2023}, pages 4531--4539.

\bibitem[{Hawkins et~al.(2020)Hawkins, Yamakoshi, Griffiths, and Goldberg}]{hawkins2020investigating}
Robert Hawkins, Takateru Yamakoshi, Thomas Griffiths, and Adele Goldberg. 2020.
\newblock \href {https://doi.org/10.18653/v1/2020.emnlp-main.376} {Investigating representations of verb bias in neural language models}.
\newblock In \emph{Proceedings of the 2020 Conference on Empirical Methods in Natural Language Processing (EMNLP)}, pages 4653--4663, Online. Association for Computational Linguistics.

\bibitem[{Hendel et~al.(2023)Hendel, Geva, and Globerson}]{hendel2023context}
Roee Hendel, Mor Geva, and Amir Globerson. 2023.
\newblock In-context learning creates task vectors.
\newblock In \emph{Findings of the Association for Computational Linguistics: EMNLP 2023}, pages 9318--9333.

\bibitem[{Jaeger and Snider(2008)}]{jaeger2008implicit}
T.~Florian Jaeger and Neal Snider. 2008.
\newblock Implicit learning and syntactic persistence: Surprisal and cumulativity.
\newblock In \emph{Proceedings of the 30th Annual Conference of the Cognitive Science Society}, volume 827812. Cognitive Science Society Austin, TX.

\bibitem[{Jaeger and Snider(2013)}]{jaeger2013alignment}
T.~Florian Jaeger and Neal~E. Snider. 2013.
\newblock Alignment as a consequence of expectation adaptation: Syntactic priming is affected by the prime’s prediction error given both prior and recent experience.
\newblock \emph{Cognition}, 127(1):57--83.

\bibitem[{Kaschak et~al.(2011)Kaschak, Kutta, and Jones}]{kaschak2011structural}
Michael~P. Kaschak, Timothy~J. Kutta, and John~L. Jones. 2011.
\newblock Structural priming as implicit learning: Cumulative priming effects and individual differences.
\newblock \emph{Psychonomic Bulletin \& Review}, 18:1133--1139.

\bibitem[{Lepori et~al.(2026)Lepori, Linzen, Yuan, and Filippova}]{lepori2026language}
Michael~A Lepori, Tal Linzen, Ann Yuan, and Katja Filippova. 2026.
\newblock Language {Models} {Struggle} to {Use} {Representations} {Learned} {In}-{Context}.
\newblock \emph{arXiv preprint arXiv:2602.04212}.

\bibitem[{Mueller et~al.(2024)Mueller, Brinkmann, Li, Marks, Pal, Prakash, Rager, Sankaranarayanan, Sharma, Sun et~al.}]{mueller2024quest}
Aaron Mueller, Jannik Brinkmann, Millicent Li, Samuel Marks, Koyena Pal, Nikhil Prakash, Can Rager, Aruna Sankaranarayanan, Arnab~Sen Sharma, Jiuding Sun, and 1 others. 2024.
\newblock The quest for the right mediator: A history, survey, and theoretical grounding of causal interpretability.
\newblock \emph{arXiv preprint arXiv:2408.01416}.

\bibitem[{Ozaki et~al.(2025)Ozaki, Bhatt, and Dillon}]{ozaki2025lstm}
Satoru Ozaki, Rajesh Bhatt, and Brian Dillon. 2025.
\newblock A {LSTM} language model learns {Hindi}-{Urdu} case-agreement interactions, and has a linear encoding of case.
\newblock \emph{Society for Computation in Linguistics}, 8(1).

\bibitem[{Pickering and Branigan(1998)}]{pickering1998representation}
Martin~J. Pickering and Holly~P. Branigan. 1998.
\newblock The representation of verbs: Evidence from syntactic priming in language production.
\newblock \emph{Journal of Memory and Language}, 39(4):633--651.

\bibitem[{Ravfogel et~al.(2020)Ravfogel, Elazar, Gonen, Twiton, and Goldberg}]{ravfogel2020null}
Shauli Ravfogel, Yanai Elazar, Hila Gonen, Michael Twiton, and Yoav Goldberg. 2020.
\newblock Null it out: Guarding protected attributes by iterative nullspace projection.
\newblock In \emph{Proceedings of the 58th annual meeting of the association for computational linguistics}, pages 7237--7256.

\bibitem[{Ravfogel et~al.(2021)Ravfogel, Prasad, Linzen, and Goldberg}]{ravfogel2021counterfactual}
Shauli Ravfogel, Grusha Prasad, Tal Linzen, and Yoav Goldberg. 2021.
\newblock Counterfactual interventions reveal the causal effect of relative clause representations on agreement prediction.
\newblock In \emph{Proceedings of the 25th Conference on Computational Natural Language Learning}, pages 194--209.

\bibitem[{Sinclair et~al.(2022)Sinclair, Jumelet, Zuidema, and Fern{\'a}ndez}]{sinclair2022structural}
Arabella Sinclair, Jaap Jumelet, Willem Zuidema, and Raquel Fern{\'a}ndez. 2022.
\newblock Structural persistence in language models: {Priming} as a window into abstract language representations.
\newblock \emph{Transactions of the Association for Computational Linguistics}, 10:1031--1050.

\bibitem[{Tang et~al.(2026)Tang, Jiang, Karray, and Hu}]{tang2026context}
Wei Tang, Xinyan Jiang, Fakhri Karray, and Lijie Hu. 2026.
\newblock In-{Context} {Learning} {Operates} as {Concept} {Subspace} {Learning}.
\newblock \emph{arXiv preprint arXiv:2605.18830}.

\bibitem[{Todd et~al.(2024)Todd, Li, Sen~Sharma, Mueller, Wallace, and Bau}]{todd2024function}
Eric Todd, Millicent Li, Arnab Sen~Sharma, Aaron Mueller, Byron Wallace, and David Bau. 2024.
\newblock Function vectors in large language models.
\newblock In \emph{International conference on learning representations}, volume 2024, pages 17282--17333.

\bibitem[{Tooley and Traxler(2010)}]{tooley2010syntactic}
Kristen~M. Tooley and Matthew~J. Traxler. 2010.
\newblock Syntactic priming effects in comprehension: A critical review.
\newblock \emph{Language and Linguistics Compass}, 4(10):925--937.

\bibitem[{Touvron et~al.(2023)Touvron, Martin, Stone, Albert, Almahairi, Babaei, Bashlykov, Batra, Bhargava, Bhosale, Bikel, Blecher, Ferrer, Chen, Cucurull, Esiobu, Fernandes, Fu, Fu, Fuller, Gao, Goswami, Goyal, Hartshorn, Hosseini, Hou, Inan, Kardas, Kerkez, Khabsa, Kloumann, Korenev, Koura, Lachaux, Lavril, Lee, Liskovich, Lu, Mao, Martinet, Mihaylov, Mishra, Molybog, Nie, Poulton, Reizenstein, Rungta, Saladi, Schelten, Silva, Smith, Subramanian, Tan, Tang, Taylor, Williams, Kuan, Xu, Yan, Zarov, Zhang, Fan, Kambadur, Narang, Rodriguez, Stojnic, Edunov, and Scialom}]{touvron2023llama}
Hugo Touvron, Louis Martin, Kevin Stone, Peter Albert, Amjad Almahairi, Yasmine Babaei, Nikolay Bashlykov, Soumya Batra, Prajjwal Bhargava, Shruti Bhosale, Dan Bikel, Lukas Blecher, Cristian~Canton Ferrer, Moya Chen, Guillem Cucurull, David Esiobu, Jude Fernandes, Jeremy Fu, Wenyin Fu, and 49 others. 2023.
\newblock \href {https://arxiv.org/abs/2307.09288} {Llama 2: Open {Foundation} and {Fine}-{Tuned} {Chat} {Models}}.
\newblock \emph{Preprint}, arXiv:2307.09288.

\bibitem[{Von~Oswald et~al.(2023)Von~Oswald, Niklasson, Randazzo, Sacramento, Mordvintsev, Zhmoginov, and Vladymyrov}]{pmlr-v202-von-oswald23a}
Johannes Von~Oswald, Eyvind Niklasson, Ettore Randazzo, Joao Sacramento, Alexander Mordvintsev, Andrey Zhmoginov, and Max Vladymyrov. 2023.
\newblock Transformers {Learn} {In}-{Context} by {Gradient} {Descent}.
\newblock In \emph{Proc. MLR}, volume 202, pages 35151--35174. PMLR.

\bibitem[{Xie et~al.(2021)Xie, Raghunathan, Liang, and Ma}]{xie2021explanation}
Sang~Michael Xie, Aditi Raghunathan, Percy~S. Liang, and Tengyu Ma. 2021.
\newblock An explanation of in-context learning as implicit bayesian inference.
\newblock \emph{arXiv:2111.02080}.

\bibitem[{Zhou et~al.(2025)Zhou, Frank, and McCoy}]{zhou-etal-2025-context}
Zhenghao Zhou, Robert Frank, and R.~Thomas McCoy. 2025.
\newblock \href {https://doi.org/10.18653/v1/2025.naacl-long.586} {Is {In}-{Context} {Learning} a {Type} of {Error}-{Driven} {Learning}? {Evidence} from the {Inverse} {Frequency} {Effect} in {Structural} {Priming}}.
\newblock In \emph{Proceedings of the 2025 Conference of the Nations of the Americas Chapter of the Association for Computational Linguistics: Human Language Technologies (Volume 1: Long Papers)}, pages 11712--11725, Albuquerque, New Mexico. Association for Computational Linguistics.

\end{thebibliography}

\appendix

\section{Relation to Linear Concept Erasure Methods}
\label{app:leace}

\paragraph{Erasure and Editing}
The continuous variable intervention method we present in Section~\ref{sec:method} is related to linear concept erasure methods such as iterative nullspace projection (INLP, for discrete concept erasure~\citealp{ravfogel2020null}) and LEAst-squares Concept Erasure (LEACE,~\citealp{belrose2023leace}), but differs in objectives. 
Concept erasure asks how to transform a representation so that a target variable is no longer linearly recoverable. 
That is, given a hidden state $h$ with target variable value $y$, erasure computes a transformed representation $h'$ such that $\text{Cov}(h', y) = 0$, such that no linear classifier or regressor can recover the target information from $h'$.
Our goal is not to remove information about a variable, but to \textit{set} that variable to arbitrary counterfactual values. 
This makes the intervention closer to targeted representation \textit{editing} than to concept \textit{erasure}. 
Erasure methods are useful for testing whether a variable is necessary; our method tests whether moving a representation along the direction associated with a continuous variable produces predictable causal effects. 
In this sense, the two approaches are complementary.

\paragraph{Shared Ideas}
Our procedure shares with LEACE the idea that a fitted linear readout provides a local coordinate system for manipulating information in representation space. 
That is, both methods rely on regression to localize the direction that correlates with the target varbale.
We also uses a normalization step analogous to the \textit{whitening} step in LEACE: by standardizing the regression coordinates, the edit is computed in a space where dimensions are isotropic and more comparable, so that the minimal-norm update is not dominated by raw variance differences across dimensions. 

\paragraph{Design Choice on PCA}
Unlike LEACE, however, we additionally perform dimensionality reduction with PCA before fitting the readout. 
This step is not conceptually required for targeted editing, as one could fit a linear readout directly in the original activation space and edit along the resulting coefficient vector. 
But it is useful in our setting for the following reasons:
\begin{enumerate}
    \item In many representation intervention settings, the dimensionality of the activation space is large while the number of distinct label values is comparatively small. In our case, we have 2200 steering vectors but only 22 distinct verb bias values.Direct regression in the full activation space can therefore identify unstable directions that fit idiosyncratic variation in the sample. By restricting the readout and edit to high-variance PCA dimensions, we bias the intervention toward directions that reflect natural variation in the vector set.

    \item PCA serves as a regularizer, restricting the readout and intervention to high-variance directions of natural variation in the vector set. This is helpful when the vectors are constructed so that salient experimental factors produce systematic variation, so that top PCA dimensions may capture major axes of this variation. Editing in PCA space therefore constrains the intervention to a low-dimensional manifold of naturally occurring representations, rather than allowing arbitrary dense perturbations in the full activation space. We do not assume that all relevant information must lie in the top PCs, but this restriction makes the intervention more conservative and easier to analyze. 

    \item The set of editable dimensions $S$ can be chosen in different ways. One can edit only the single most predictive PC, the top $k$ ranked PCs, PCs selected by Lasso regularization, or all PCA dimensions. Importantly, selecting multiple PCs does not mean that each individual edit moves the representation in many independent directions. For a fixed target value, the actual update is still a single vector, $\Delta \tilde{z}_i \propto \beta_S$, lying inside the selected subspace. Increasing the number of editable PCs changes the subspace in which this one-dimensional update direction is allowed to lie. Thus, the choice of $S$ controls the representational basis available to the intervention, while the counterfactual edit for each vector remains a directed movement toward the chosen target value.
\end{enumerate}

\begin{figure}[h]
    \centering
    \includegraphics[width=0.95\columnwidth]{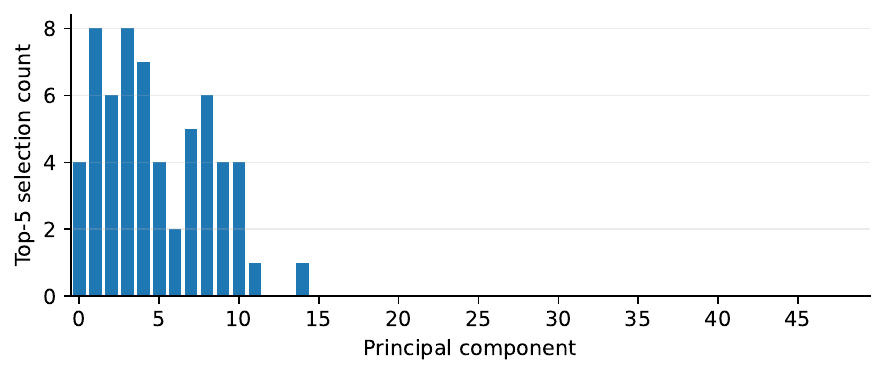}
    \caption{The frequency of each of the 50 principle components to be among the top 5 selected editable PCs across all experiments on verb bias intervention. Most selected PCs are concentrated in the top 10 PCs that explain the most variance in the steering vectors.}
    \label{fig:app_topPC_freq}
\end{figure}

In this paper, we reported results for editing the top 5 PCs sorted by the absolute values of their regression coefficients. We report the frequency of each of the 50 PCs that is selected to be edited across experiments (i.e., the frequency of a PC to be among the top 5 PCs) in Figure~\ref{fig:app_topPC_freq}. It shows that most of the edited PCs are among the first 10 PCs that explains the most variance. This empirical observation confirms the design choices we made, such that in the steering vectors we constructed, the verb bias information is indeed captured by major axes of variance.
We leave it to future studies on the effect of PCA for other variables, as well as the effect of different editable dimensions.

\section{Diagnostic Probes for Verb Bias, Prime Structure, and Error Signal in Steering Vectors}
\label{app:diagnostics}

\paragraph{Motivation}
A potential concern in our intervention experiments is that the regression model may exploit idiosyncratic information about the prime verb rather than recover a generalizable representational dimension corresponding to the target variable. 
This concern is especially important in our setting because each target value is defined at the verb level: all steering vectors extracted from the same prime verb share the same verb bias label value for regression, and all steering vectors extracted from the same verb--structure combination share the same signed-error value. 
We therefore conduct leave-one-verb-out diagnostic analyses to test whether the relevant variables generalize to held-out verbs rather than merely separating repeated samples of the same verb.

\begin{figure}[h]
    \centering
    \includegraphics[width=0.9\columnwidth]{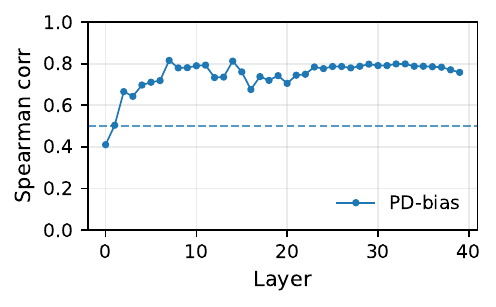}
    \caption{Leave-one-verb-out diagnostic for PD verb bias. We report Spearman correlation between true and predicted held-out verb bias across layers.}
    \label{fig:app_diag_bias}
\end{figure}

\begin{figure}[h]
    \centering
    \includegraphics[width=0.9\columnwidth]{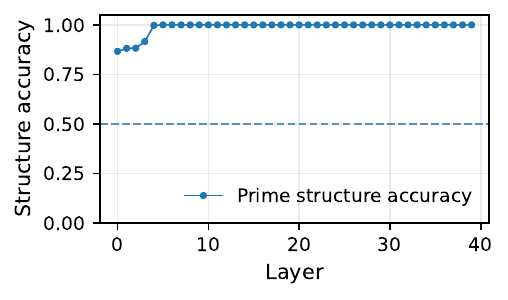}
    \caption{Leave-one-verb-out diagnostic for prime structure. We report held-out accuracy for classifying whether a steering vector was extracted from a DO-prime or PD-prime context.}
    \label{fig:app_diag_structure}
\end{figure}

\begin{figure}[h]
    \centering
    \includegraphics[width=0.9\columnwidth]{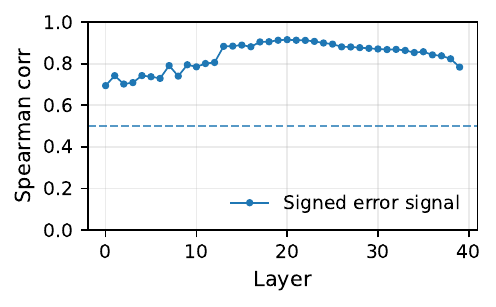}
    \caption{Leave-one-verb-out diagnostic for signed error signal. We report Spearman correlation between true and predicted held-out error signal values across layers.}
    \label{fig:app_diag_error}
\end{figure}

\paragraph{Procedure}
For each layer, we first collect steering vectors from all prime verbs and both prime structures. 
We then fit a diagnostic model on vectors from 21 verbs and evaluate on the held-out verb, repeating this procedure for all verbs. 
For continuous variables such as PD bias and error signal, we compute the mean predicted value and the true value for each held-out verb, and report the Spearman correlation across held-out verbs. 
This evaluates whether the probe recovers the correct ordering of verbs, rather than only minimizing sample-level error. 
For the binary prime structure variable, we fit a binary classifier to distinguish DO-prime from PD-prime steering vectors and report held-out classification accuracy.

\paragraph{Results}
Figure~\ref{fig:app_diag_bias} shows the diagnostic result for PD verb bias. 
The correlation is weak in early layers (layer 0 through 7) but rises substantially to 0.8 around layer 10 and become stable in middle and later layers (after layer 23), indicating that the static verb bias information is not merely memorized at the sample level but is recoverable in a way that generalizes across verbs. 
This supports the interpretation of Experiment~2: the intervention direction targets a graded lexical-bias dimension that is present in the steering vector space.

Figure~\ref{fig:app_diag_structure} shows the diagnostic result for prime structure. Prime structure is decoded with near-ceiling accuracy from relatively early layers onward (starting at layer 5). 
This indicates that the vectors reliably encode whether the prime context contains a DO or PD structure. 
Importantly, this does not by itself imply that prime-structure information is already \textit{integrated} with verb bias information in a way that causally supports downstream adaptation.
As discussed in Section~\ref{sec:exp3}, error signal intervention becomes effective only in middle-to-later layers.

Figure~\ref{fig:app_diag_error} shows the diagnostic result for error signal. 
We define signed error as the difference between the observed prime structure and the verb's static PD bias, so that positive values correspond to a DO-directed update and negative values correspond to a PD-directed update. 
Among several theory-motivated variants that we tested, including absolute error, squared error, surprisal, and signed surprisal, this signed error target produced the strongest overall diagnostic performance. 
The correlation gradually rises in early layers and peaks at 0.91 in middle layers (layer 17 through 23).
\section{Layerwise Dynamics}
\label{app:layerwise}

\subsection{Additional Plots for Experiment 2}

\begin{figure*}[h]
    \centering
    \includegraphics[width=0.8\textwidth]{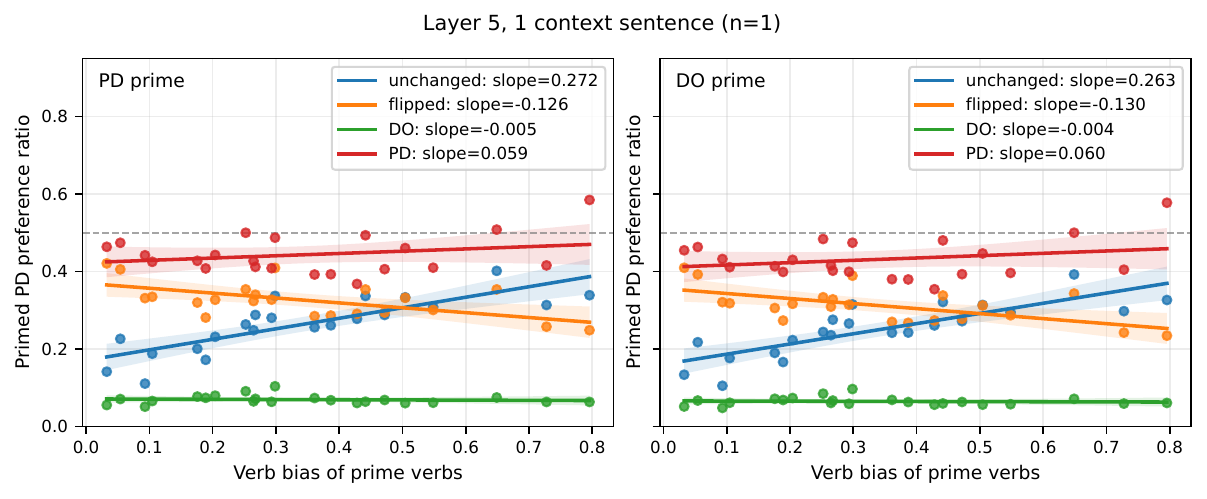}
    \vspace{1ex}
    \includegraphics[width=0.8\textwidth]{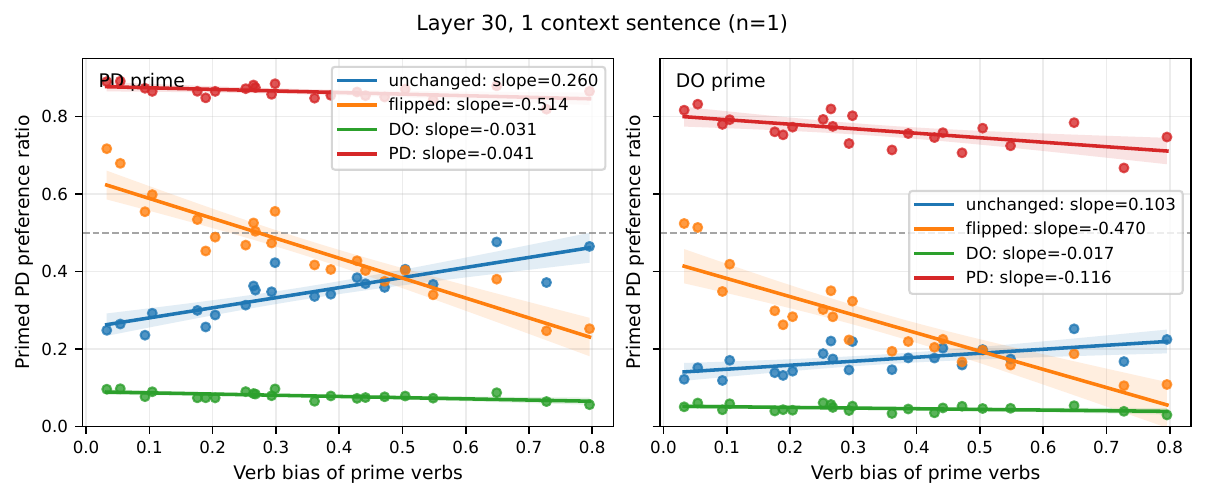}
    \vspace{1ex}
    \includegraphics[width=0.8\textwidth]{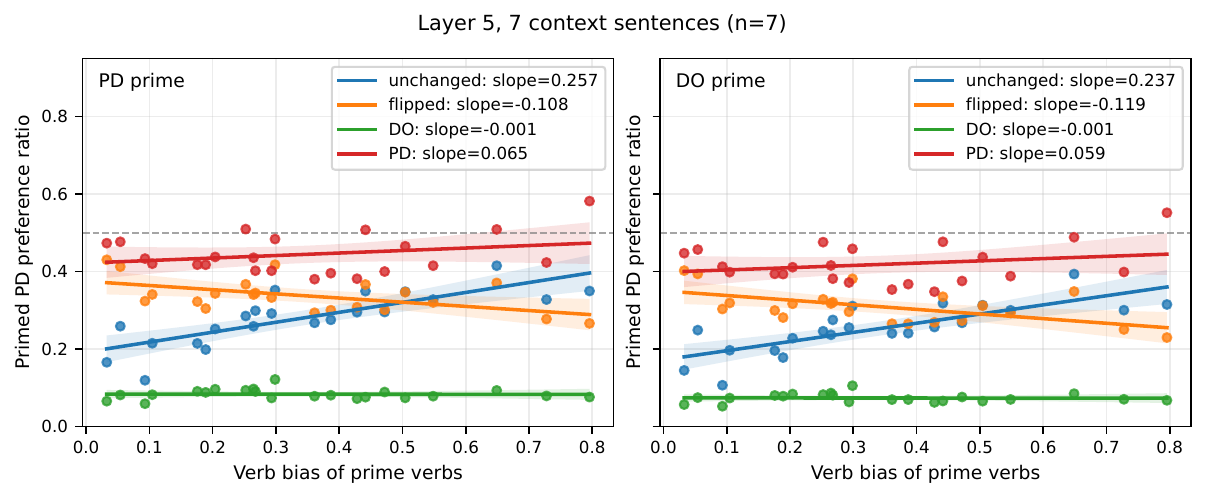}
    \vspace{1ex}
    \includegraphics[width=0.8\textwidth]{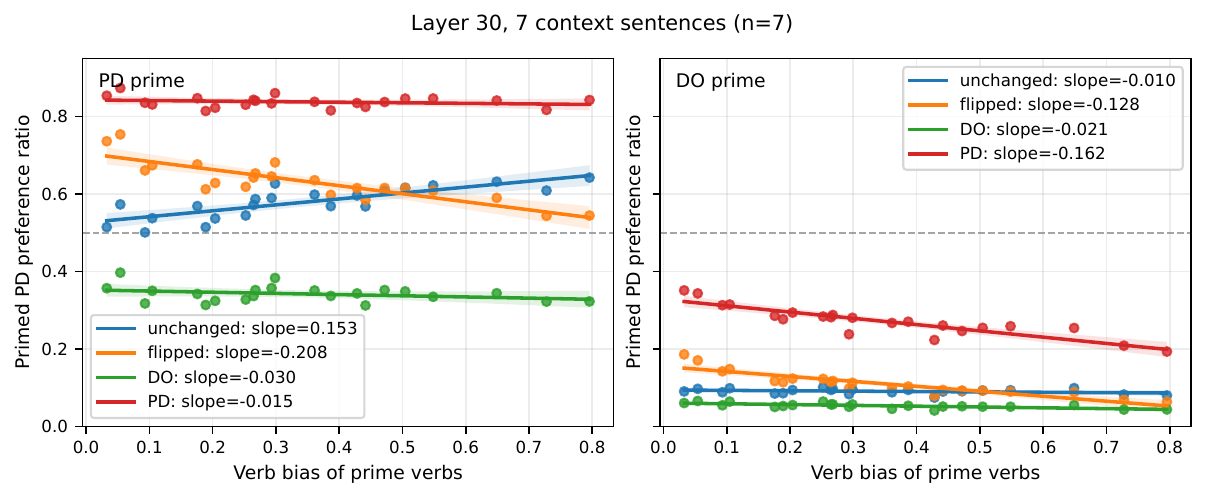}
    \caption{Additional results for Experiment 2: downstream structural preference changes as results of counterfactual intervention on verb bias in steering vectors extracted from layer 5 and 30, with 1 or 7 context sentences.}
    \label{fig:app_c_exp2}
\end{figure*}

Figure~\ref{fig:app_c_exp2} provides additional layerwise results for the verb bias intervention in Experiment 2. 
The same qualitative pattern emerges already at layer 5: editing the steering vector toward the PD-bias value increases the target sentence's PD preference, editing toward the DO-bias value decreases it, and the flipped condition falls between these two endpoints. 
This suggests that the static verb bias component of the steering vector is already causally accessible in relatively early layers. 
However, the separation between edit conditions becomes more systematic in later layers, as shown by the layer-30 plots, where the PD and DO edit conditions form clearer and more stable offsets from the unchanged condition. 

Comparing $n=1$ and $n=7$ (number of context sentences, or primes, in an ICL sequence from which the steering vectors are extracted) further suggests that longer priming contexts amplify the structural component of the steering vector: with more prepended context sentences, the intercept differences between PD-prime and DO-prime conditions become larger, indicating a stronger overall pull toward the prime structure. 
Importantly, this does not eliminate the effect of static verb bias: within each prime condition, the fitted lines continue to vary with the prime verb's bias. Thus, the static-bias intervention effect is robust across layers and context lengths, while the relative contribution of prime-structure information becomes more dominant when the priming context is stronger.

\begin{figure}[h]
    \centering
    \includegraphics[width=0.48\textwidth]{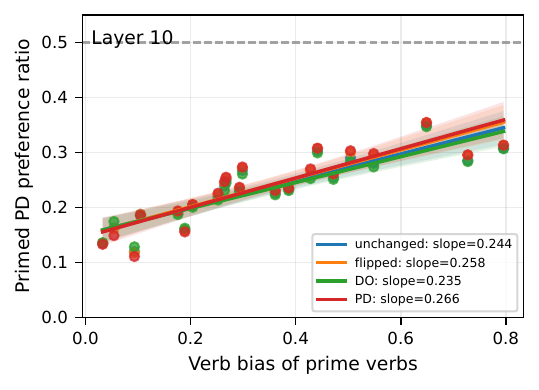}
    \vspace{1ex}
    \includegraphics[width=0.48\textwidth]{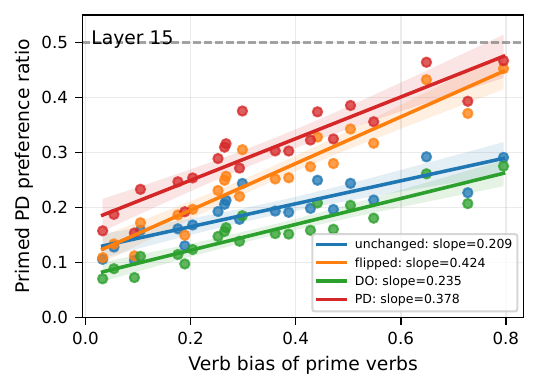}
    \caption{Counterfactual intervention on error signals for layer 10 (top) and 15 (bottom). Our counterfactual editing has no downstream effects on error signals in earlier layers (for example, layer 10) but becomes effective starting at layer 15.}
    \label{fig:app_c_exp3}
\end{figure}

\subsection{Additional Plots for Experiment 3}

We further examined whether the causal effect of error signal editing is already present in earlier layers.
Figure~\ref{fig:app_c_exp3} compares the raw, non-delta intervention plots for the DO-prime condition at layers 10 and 15. 
At layer 10, all four editing conditions largely overlap: the \texttt{Unchanged}, \texttt{Flip}, \texttt{DO-biased}, and \texttt{PD-biased} steering vectors produce very similar fitted slopes and intercepts, indicating that changing the signed-error value in the vector does not yet systematically affect downstream target preferences. 
By contrast, at layer 15, the intervention conditions separate clearly. Editing toward the PD signed-error value increases the primed PD preference most strongly, the flipped condition also increases the slope relative to unchanged, and the DO-edited condition remains closest to or below unchanged. 
This contrast supports the claim that signed-error information is not causally organized in the steering vectors from the earliest layers, even though diagnostic probes can already decode prime-structure information earlier. 
Rather, the causal effect of signed-error editing emerges only after the layer range where prime structure and verb-bias information appear to be integrated into a direction that can influence downstream structural preference.

\section{An Elaboration on the Error Signal Counterfactual Editing}
\label{app:error_signal}

\begin{figure}[h]
  \centering
  \includegraphics[width=0.9\columnwidth]{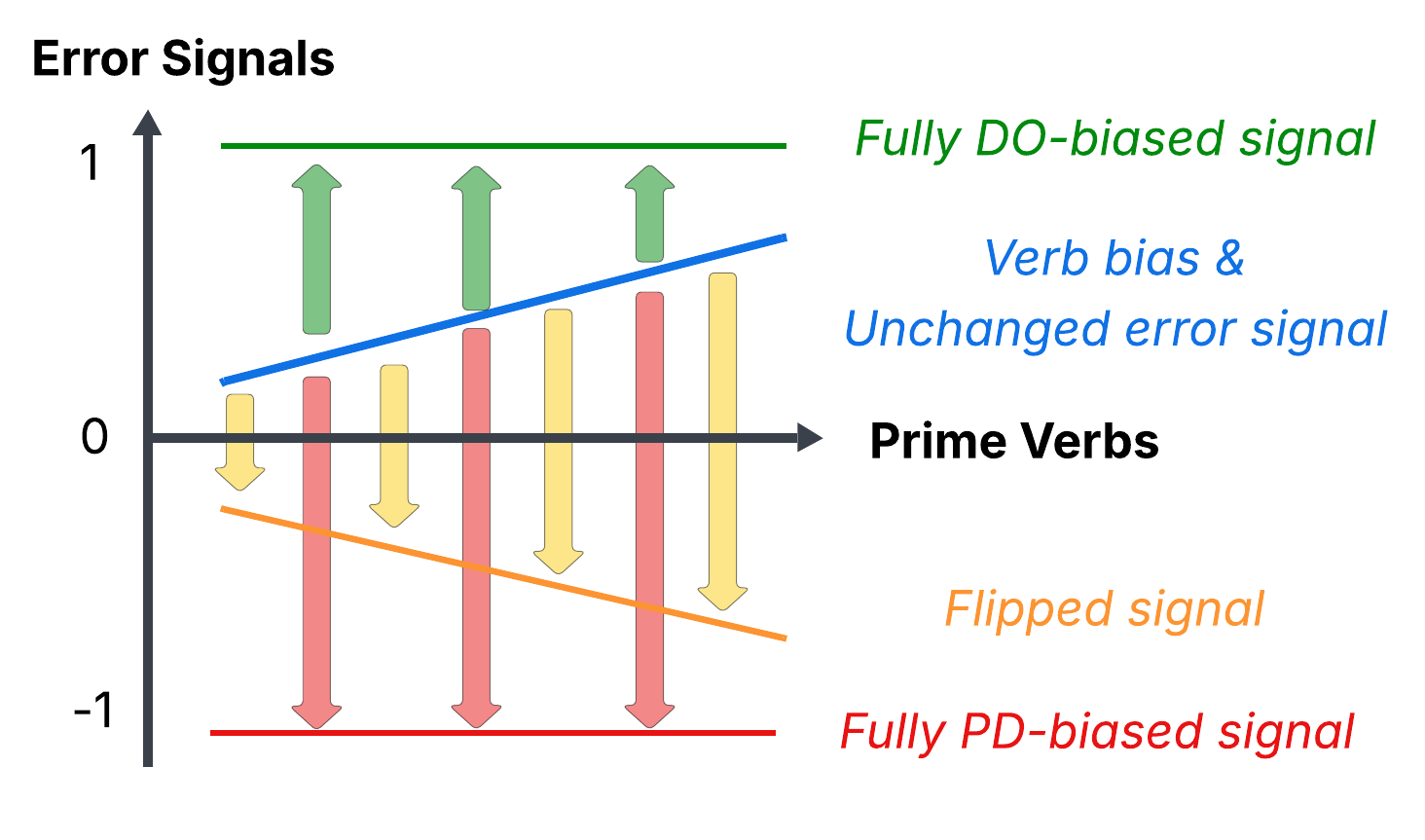}  
  \caption{A demonstration of counterfactual editing on the signed error signal variable.}
  \label{fig:app_error_signal_demo}
\end{figure}

We elaborate on the logic of the predictions on the slopes and intercepts of $\Delta \text{Pref}_{\text{edited}}$ for the three counterfactual editing conditions in Experiment 3.
For a given prime verb $v$, let $b_v \in (0,1)$ denote its PD bias, where larger values indicate stronger baseline preference for the PD construction. In the error signal analysis, the relevant quantity is the \textit{mismatch} between the structure observed in the prime and the structure expected from the verb's static bias. 
Intuitively, a DO prime with a highly PD-biased verb is more surprising than a DO prime with a DO-biased verb, and a PD prime with a highly DO-biased verb is more surprising than a PD prime with a PD-biased verb. 
Thus, unlike the static verb bias intervention, the signed-error intervention is predicted to produce effects that depend jointly on prime structure and verb bias.

In Experiment 3, we measure the effect of each edit relative to the unchanged steering vector. For each prime verb $v$ and prime structure $s$, we define
$\Delta \text{Pref}_{\text{edited}}(v,s)=\text{Pref}_{\text{edited}}(v,s) -\text{Pref}_{\text{unchanged}}(v,s)$,
where $\text{Pref}$ is the model's PD preference ratio (Equation~\ref{eq:prime_pref}) on the target sentences after steering vector injection. 
This subtraction removes the baseline priming effect already carried by the unedited steering vector. 
Therefore, the delta metric asks: after holding fixed the effect of original steering vectors, how much does changing the error signal direction \textit{additionally} move the model's downstream target preference?
A positive slope with the delta metric means that the edit has a larger PD-increasing effect for more PD-biased prime verbs; a negative slope means that the edit has a larger PD-increasing effect for more DO-biased prime verbs. 
The left edge of each fitted line can be interpreted as the approximate effect for strongly DO-biased verbs, while the right edge reflects the effect for strongly PD-biased verbs.

Consider the DO-prime condition. A positive error signal means updating towards the DO direction.
As is illustrated by the blue slope in Figure~\ref{fig:app_error_signal_demo}, the signed error is small for DO-biased verbs since the observed structure is consistent with the verb's baseline preference, while it is large for PD-biased verbs, because a DO prime is unexpected given the verb's strong PD bias. 
Therefore, if the signed-error direction is causally used as a counter-bias update signal, increasing the PD-oriented signed-error component should have a larger PD-increasing effect for PD-biased prime verbs than for DO-biased prime verbs. This predicts a \textit{positive} slope for edits that increase the PD-oriented signed-error component.

The three edit conditions differ in how they change this component:
\begin{enumerate*}[label=(\arabic*)]
    \item The \texttt{PD} edit (illustrated by the red slope in Figure~\ref{fig:app_error_signal_demo}) sets the vector toward a uniformly PD-oriented error state. This should increase PD preference across the full range of prime verbs, yielding a positive intercept and a positive slope: $\mathrm{slope}(\Delta \mathrm{Pref}_{\mathrm{PD}}) > 0$ and $\mathrm{intercept}(\Delta \mathrm{Pref}_{\mathrm{PD}}) > 0$.
    
    \item The \texttt{{flip}} edit (illustrated by the orange slope in Figure~\ref{fig:app_error_signal_demo}) reverses the signed-error value associated with each prime verb. Under DO primes, this has little effect for very DO-biased verbs, whose original signed error is already near the DO-consistent end of the scale, but it has a large effect for highly PD-biased verbs, whose original signed error is strongly positive and is therefore strongly altered by the flip. Thus, the flipped condition is predicted to start near zero but increase sharply with static PD bias: $\mathrm{intercept}(\Delta \mathrm{Pref}_{\mathrm{flip}}) \approx 0$ and $\mathrm{slope}(\Delta \mathrm{Pref}_{\mathrm{flip}}) > 0$.

    \item The \texttt{DO} edit pushes the vector toward a DO-oriented state. Under DO primes, this is largely congruent with the observed structure and therefore should not produce a strong counter-bias update. Moreover, because the model's PD preference is already low in many DO-prime cases, the possible downward movement is limited by a floor effect. We therefore expect the DO edit to have a small effect, with an intercept below zero and a weak positive slope:$\mathrm{intercept}(\Delta \mathrm{Pref}_{\mathrm{DO}}) < 0$ and $\mathrm{slope}(\Delta \mathrm{Pref}_{\mathrm{DO}}) > 0$ but small.

\end{enumerate*}

Because the flipped edit preserves the verb-specific ordering of error signal magnitudes while reversing their orientation, it is expected to produce the strongest dependence on verb bias. 
This leads to the following prediction:
$\mathrm{slope}(\Delta \mathrm{Pref}_{\mathrm{flip}}) > \mathrm{slope}(\Delta \mathrm{Pref}_{\mathrm{PD}})  > \mathrm{slope}(\Delta \mathrm{Pref}_{\mathrm{DO}})$.
Putting these together, the qualitative prediction for DO primes is:
$
\mathrm{slope}(\Delta \mathrm{Pref}_{\mathrm{flip}})
>
\mathrm{slope}(\Delta \mathrm{Pref}_{\mathrm{PD}})
>
\mathrm{slope}(\Delta \mathrm{Pref}_{\mathrm{DO}})
> 0
$
and
$
\mathrm{intercept}(\Delta \mathrm{Pref}_{\mathrm{PD}})
>
\mathrm{intercept}(\Delta \mathrm{Pref}_{\mathrm{flip}})
\approx 0
>
\mathrm{intercept}(\Delta \mathrm{Pref}_{\mathrm{DO}}).
$
The same reasoning can be applied to PD prime.


\section{An Elaboration on Psycholinguistic Theories of Structural Priming}
\label{app:priming_theory}

\begin{figure}[h]
  \centering
  \includegraphics[width=\columnwidth]{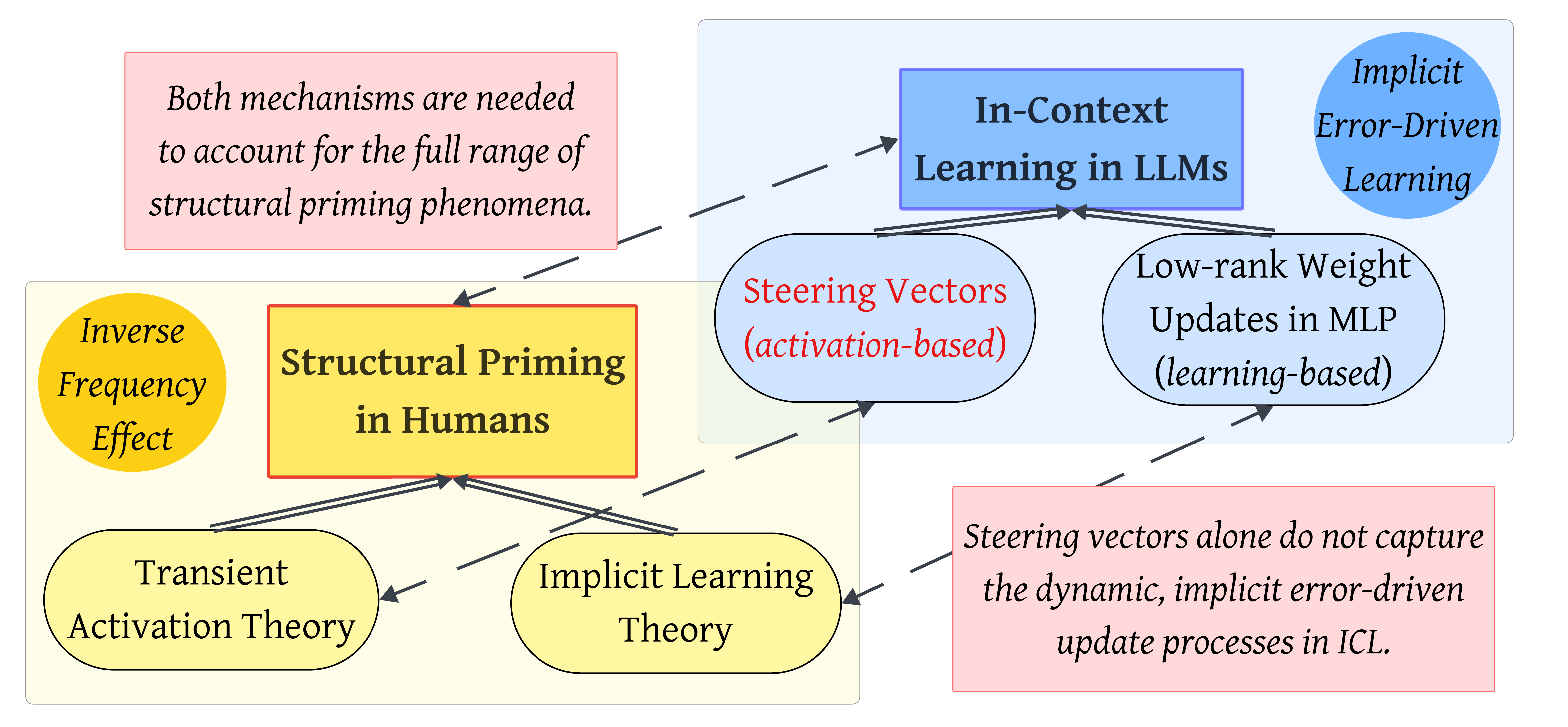}  
  \caption{A mapping between human psycholinguistic theories of structural priming and ingredients of ICL.}
  \label{fig:implication}
\end{figure}

Structural priming has traditionally been explained by two broad classes of theories.
Under the transient activation account \cite{pickering1998representation}, processing a prime sentence temporarily activates the corresponding structural representation, making the same structure easier to retrieve or reuse in a subsequent target sentence. 
This account explains why a recently encountered structure can increase the probability of producing or comprehending the same structure again through an activation perspective.
However, in its standard form, transient activation does not predict the inverse frequency effect (IFE): the residual activation of a structure is not expected to depend on how surprising the prime structure was given the lexical bias of the prime verb. 

By contrast, the implicit learning account \cite{chang2006becoming} treats structural priming as a consequence of experience-driven updates to the learner's probabilistic expectations about grammatical structures, including verb-specific biases. 
On this view, language users continuously use prior experience to predict upcoming linguistic input, and update their expectations when those predictions are violated. 
Crucially, because the update is error-driven, a less expected prime structure should induce a larger update than a more expected one. 
In the case of dative alternation, a PD prime with a strongly DO-biased verb, or a DO prime with a strongly PD-biased verb, should therefore produce a stronger priming effect. 
This prediction is the inverse frequency effect. 

The two accounts are not mutually exclusive: under a dual-mechanism view \cite{tooley2010syntactic}, structural priming may reflect both short-lived activation of recently processed structures and longer-term error-driven adaptation.

\section{Computational Resource, Package, and AI Usage}
All experiments reported in this paper were conducted on the computing cluster of the authors' affiliated institution, using NVIDIA H200 Tensor Core GPUs (141GB). Causal interventions and behavioral evaluations took approximately 80 GPU hours.
All regression experiments, PCA dimensionality reduction, and normalization were conducted with the \texttt{scikit-learn v1.7.1} packgage.
We used ChatGPT 5.5 and Copilot for coding and plotting.

\end{document}